**ORIGINAL PAPER**

# Artificial virtuous agents in a multi-agent tragedy of the commons

Jakob Stenseke[1]

**Abstract**
Although virtue ethics has repeatedly been proposed as a suitable framework for the development of artificial moral agents (AMAs), it has been proven difficult to approach from a computational perspective. In this work, we present the first technical implementation of artificial virtuous agents (AVAs) in moral simulations. First, we review previous conceptual and technical work in artificial virtue ethics and describe a functionalistic path to AVAs based on dispositional virtues, bottom-up learning, and top-down eudaimonic reward. We then provide the details of a technical implementation in a moral simulation based on a tragedy of the commons scenario. The experimental results show how the AVAs learn to tackle cooperation problems while exhibiting core features of their theoretical counterpart, including moral character, dispositional virtues, learning from experience, and the pursuit of eudaimonia. Ultimately, we argue that virtue ethics provides a compelling path toward morally excellent machines and that our work provides an important starting point for such endeavors.

## 1 Introduction

Over the last decades, the rapid development and application of artificial intelligence (AI) has spawned a lot of research focusing on various ethical aspects of AI (AI ethics), and the prospects of implementing ethics into machines (machine ethics)[1]. The latter project can further be divided into theoretical debates on machine morality[2], conceptual work on hypothetical artificial moral agents (Malle 2016), and more technically oriented work on prototypical AMAs[3]. Following the third branch, the vast majority of the technical work has centered on constructing agent-based deontology (Anderson and Anderson 2008; Noothigattu et al. 2018), consequentialism (Abel et al. 2016; Armstrong 2015), or hybrids (Dehghani et al. 2008; Arkin 2007).

Virtue ethics has repeatedly been suggested as a promising blueprint for the creation of artificial moral agents (Berberich and Diepold 2018; Coleman 2001; Gamez et al. 2020; Howard and Muntean 2017; Wallach and Allen 2008; Mabaso 2020; Sullins 2021; Navon 2021; Stenseke 2021)[4]. Beyond deontological rules and consequentialist utility functions, it presents a path to construe a more comprehensive picture of what it in fact is to have a moral character and be a competent ethical decision maker in general. With the capacity to continuously learn from experience, be context-sensitive and adaptable to changes, an AMA based on virtue ethics could potentially accommodate the subtleties of human values and norms in complex and dynamic environments. However, although previous work has proposed that artificial virtue could be realized through connectionism and the recent advancements made with artificial neural networks and machine learning (Wallach and Allen 2008; Howard and Muntean 2017; Gips 1995; DeMoss 1998), hardly any technical work has attempted to do so (Tolmeijer et al. 2020). The major reason is that virtue ethics has been proven difficult to tackle from a computational point of view (Tolmeijer et al. 2020; Bauer 2020; Lindner et al. 2020; Arkin 2007). While action-centric frameworks such as consequentialism and deontology offer more or less straight-forward instructions convenient for algorithmic implementation, those who try to construct a virtuous machine quickly find themselves overwhelmed with the task of figuring out how generic virtues relate to moral behavior and how to interpret seemingly intangible concepts such as *moral character*, *eudaimonia* ("flourishing"), *phronesis* ("practical wisdom"), and *moral exemplars*. This conundrum is further illustrated in Fig. 1.

---

[1] For a broader introduction to machine ethics, see Wallach and Allen (2008), Anderson and Anderson (2011), and Pereira et al. (2016).

[4] Virtue ethics has also recently been explored in the context of social robotics and human–robot interaction (Constantinescu and Crisp 2022; Cappuccio et al. 2021; Sparrow 2021; Peeters and Haselager 2021).

[2] See Behdadi and Munthe (2020) for an excellent summary of these debates.

[3] For two recent surveys on implementations in machine ethics, see Tolmeijer et al. (2020) and Cervantes et al. (2020).

✉ Jakob Stenseke
  jakob.stenseke@fil.lu.se

1 Department of Philosophy, Lund University, Lund, Sweden

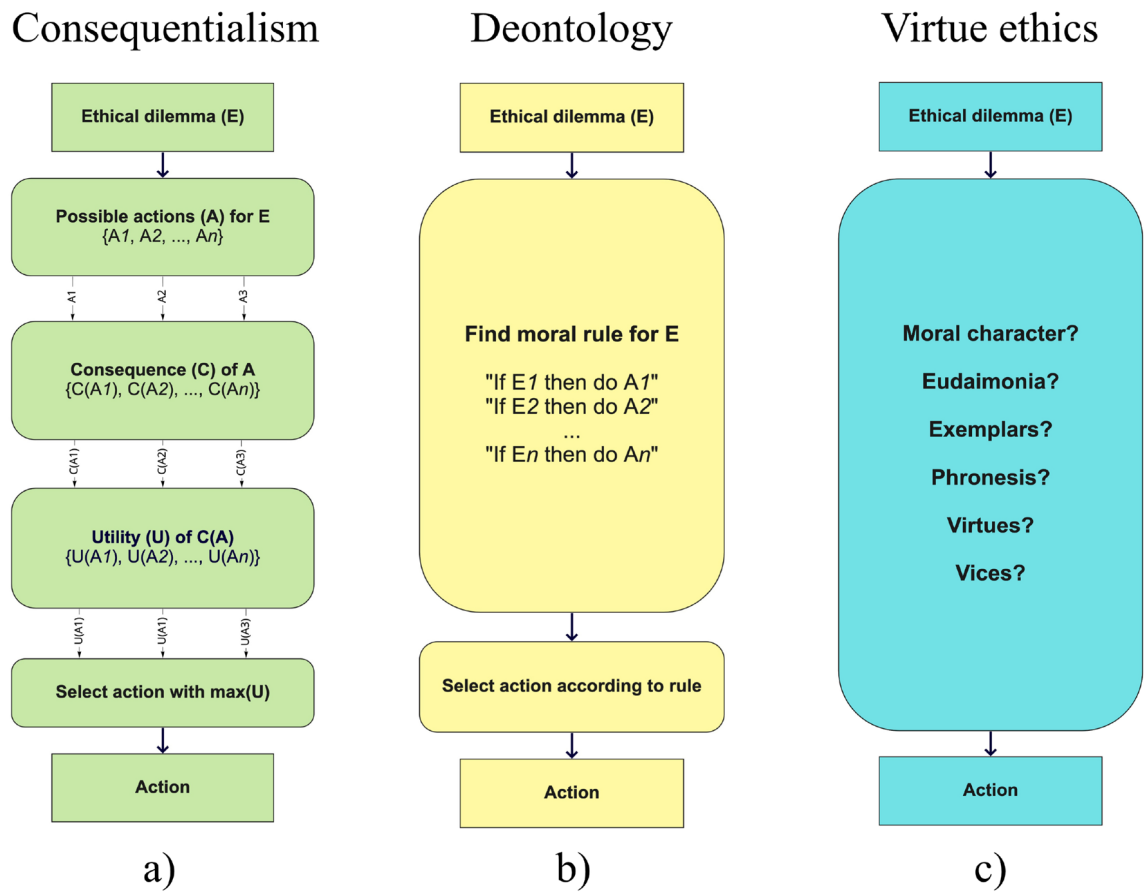

**Fig. 1** Rough sketches of three basic ethical algorithms. **a** Consequentialism: Given an ethical dilemma E, a set of possible actions in E, a way of determining the consequences of those actions and their resulting utility, the consequentialist algorithm will perform the action yielding the highest utility. **b** Deontology: Given an ethical dilemma E and set of moral rules, the deontological algorithm will search for the appropriate rule for E and perform the action dictated by the rule. **c** Virtue ethics: By contrast, constructing an algorithm based on virtue ethics presents a seemingly intriguing puzzle

In this paper, we refine and extend the technical details of a conceptual model (Stenseke 2021) and present the first experimental implementation of AMAs that solely focuses on virtue ethics. The experimental results show that our AVAs manage to tackle cooperation problems while exhibiting core features of their theoretical counterpart, including moral character, dispositional virtues, learning from experience, and the pursuit of eudaimonia. The main aim is to show how virtue ethics offers a promising framework for the development of moral machines that can be suitably incorporated in real-world domains.

The paper is structured as follows. In Sects. 1.1 and 1.2, we survey previous conceptual and technical work in artificial virtue and outline an eudaimonic version of the theory based on functionalism and connectionist learning. In Sects. 2.1, we outline the computational model of an AVA with dispositional virtues and a phronetic learning system based on eudaimonic reward. In Sects. 2.2, we introduce the ethical environment *BridgeWorld*, a virtual tragedy of the commons scenario where a population of artificial agents have to balance cooperation and self-interest in order to prosper. In Sects. 2.3, we describe how AVAs based on our computational model are implemented in the environment and provide the technical details of the experimental setup. In the remaining sections, we present the experimental results (Sects. 3), discuss a number of persisting challenges, and describe fruitful venues for future work (Sects. 4).

## 1.1 Virtue ethics

Virtue ethics refers to a large family of ethical traditions that can be traced back to Aristotle and Plato in the West and Confucius and Mencius in the East [5]. In Western moral philosophy of the modern day, it has claimed its place as one of the three central frameworks in normative ethics through the work of Anscombe (1958), Nussbaum (1988), Slote (1983, 1992), Hursthouse (1999), and Annas (2011).

Essentially, virtue ethics is about *being* rather than *doing*. Rather than looking at actions themselves (deontology) or the consequences of actions (consequentialism), the virtuous agent nurtures the character traits that allows her to *be* morally virtuous. In this way, virtues can be viewed as the morally praiseworthy dispositions—e.g., courage, temperance, fairness—that an agent has or strives to have. Central to virtue ethics is also the concept of *phronesis* ("practical wisdom"), which, according to Aristotle, can be defined as "a true and reasoned state of capacity to act with regard to the things that are good or bad for man" (NE VI.5). Not only does phronesis encompass the ability to achieve certain ends, but also to exercise good judgment in relation to more general ideas of the agents well-being. To that end, phronesis is often construed as the kind of moral wisdom gained from experience that a virtuous adult has but a nice child lacks: "[...] a young man of practical wisdom cannot be found. The case is that such wisdom is concerned not only with universals but with particulars, which become familiar from experience" (NE 1141b 10).

While most versions of virtue ethics agree on the central importance of virtue and practical wisdom, they disagree about the way these are combined and emphasized in different aspects of ethical life. For instance, eudaimonist versions of virtue ethics (Hursthouse 1999; Ryan et al. 2008) define virtues in relation to *eudaimonia* (commonly translated as "well-being" or "flourishing"), in the sense that the former (virtues) are the traits that supports an agent to achieve the latter (eudaimonia). That is, for an eudaimonist, the key reason for developing virtues is that they contribute to an agent's eudaimonia. Agent-based and exemplarist versions, on the other hand, hold that the normative value of virtues is best explained in terms of dispositions and motivations of the agent and that these qualities are most suitably characterized in moral exemplars (Slote 1995; Zagzebski 2010)[6].

## 1.2 Previous work in artificial virtue

The various versions of virtue ethics have given rise to a rather diverse set of approaches to artificial virtuous agents, ranging from narrow applications and formalizations to more general and conceptual accounts. Of the work that explicitly considers virtue ethics in the context of AMAs, it is possible to identify five prominent themes: (1) the skill metaphor developed by Annas (2011), (2) the virtue-theoretic action guidance and decision procedure described by Hursthouse (1999), (3) learning from moral exemplars, (4) connectionism about moral cognition, and (5) the emphasis on function and role.

(1) The first theme is the idea that virtuous moral competence—including actions and judgments—is acquired and refined through active *intelligent practice*, similar to how humans learn and exercise practical skills such as playing the piano (Annas 2011; Dreyfus 2004). In a machine context, this means that the development and refinement of artificial virtuous cognition ought to be based on a continuous and interactive learning process, which emphasizes the "bottom-up" nature of moral development as opposed to a "top-down" implementation of principles and rules (Howard and Muntean 2017).

(2) The second theme, following Hursthouse (1999), is that virtue ethics can provide action guidance in terms of "v-rules" that express what virtues and vices command (e.g., "do what is just" or "do not what is dishonest"), and offers a decision procedure in the sense that "An action is right iff it is what a virtuous agent would characteristically (i.e., acting in character) do in the circumstances" (Hursthouse (1999), p. 28). Hursthouse's framework has been particularly useful as a response against the claim that virtue ethics is "uncodifiable" and does not provide a straight-forward procedure or "moral code" that can be used for algorithmic implementation (Bauer 2020; Arkin 2007; Tonkens 2012; Gamez et al. 2020).

(3) The third theme is the recognition that moral exemplars provide an important source for moral education (Hursthouse 1999; Zagzebski 2010; Slote 1995). In turn, this has inspired a moral exemplar approach to artificial virtuous agents, which centers on the idea that artificial agents can become virtuous by imitating the behavior of excellent virtuous humans (Govindarajulu et al. 2019; Berberich and Diepold 2018; Mabaso 2020). Apart from offering convenient means for control and supervision, one major appeal of the approach is that it could potentially resolve the *alignment problem*, i.e., the problem of aligning machine values with human values (Armstrong 2015; Gabriel 2020).

---
[5] See Crisp and Slote (1997) and Devettere (2002) for two outstanding introductions to virtue ethics.

[6] However, this does not mean that moral exemplars are unimportant for eudaimonists, as they can serve to explain how one can, e.g., identify virtues and the aims of virtuous action (Hursthouse 1999). See Hursthouse and Pettigrove (2018) for a comprehensive description of contemporary directions in virtue ethics and their variations.

(4) The fourth theme is based on the relationship between virtue ethics and connectionism, i.e., the cognitive theory that mental phenomena can be described using artificial neural networks. The emphasis on learning, and the possibility to apprehend context-sensitive and non-symbolic information without general rules, has indeed led many authors to highlight the appeal of unifying virtue ethics with connectionism (Berberich and Diepold 2018; Wallach and Allen 2008; Howard and Muntean 2017; Stenseke 2021). The major reason is that it would provide AVAs with a compelling theoretical framework to account for the development of moral cognition (Churchland 1996; DeMoss 1998; Casebeer 2003), as well as the technological promises of modern machine learning methods (e.g., deep learning and reinforcement learning).

(5) The fifth theme is the virtue-theoretic emphasis on function and role (Coleman 2001; Thornton et al. 2016). According to both Plato (R 352) and Aristotle (NE 1097b 26-27), virtues are those qualities that enable an agent to perform their function well. The virtues of an artificial agent would, consequently, be the traits that allow it to effectively carry out its function. For instance, a self-driving truck does not share the same virtues as a social companion robot used in childcare; they serve different roles, are equipped with different functionalities, and meet their own domain-specific challenges. Situating artificial morality within a broader virtue-theoretic conception of function would therefore allow us to clearly determine the relevant traits a specific artificial agent needs in order to excel at its particular role.

The biggest challenge for the prospect of AVAs is to move from the conceptual realm of promising ideas to the level of formalism and details required for technical implementation. Guarini (2006, 2013a, 2013b) has developed neural network systems to deal with the ambiguity of moral language, and in particular the gap between generalism and particularism. Without the explicit use of principles, the neural networks can learn to classify cases as morally permissible/impermissible. Inspired by Guarini's classification approach, Howard and Muntean have broadly explored the conceptual and technical foundations of autonomous artificial moral agents (AAMAs) based on virtue ethics (Howard and Muntean 2017). Based on Annas skill metaphor (Annas 2011) and the moral functionalism of Jackson and Pettit (1995), they conjecture that artificial virtues (seen as dispositional traits) and artificial moral cognition can be developed and refined in a bottom-up process through a combination of neural networks and evolutionary computation methods. Their central idea is to evolve populations of neural networks using an evolutionary algorithm that, via a fitness selection, alters the parameter values, learning functions, and topology of the networks. The emerging candidate solution is the AAMA with "a minimal and optimal set of virtues that solves a large enough number of problems, by optimizing each of them" (Howard and Muntean (2017), p. 153). Although promising in theory, Howard and Muntean's proposed project is lacking in several regards. First, while combinations of neural networks and randomized search methods have yielded promising results in well-defined environments using NeuroEvolution of Augmenting Topologies (Stanley and Miikkulainen 2002) (NEATs) or deep reinforcement learning (Berner et al. 2019), Howard and Muntean's proposal turns into a costly search problem of infinite dimensions. Furthermore, due to the highly stochastic process of evolving neural networks and an equivocal definition of fitness evaluation, it is not guaranteed that morally excellent agents would appear even if we granted infinite computational resources. Besides being practically infeasible, several crucial details of their implementation are missing, and they only provide fragmentary results of an experiment where neural networks learn to identify anomalies in moral data. It therefore remains unclear how their envisioned AAMAs ought to be implemented in moral environments apart from the classification tasks investigated by Guarini (2006).

Berberich and Diepold (2018) have, in a similar vein, broadly described how various features of virtue ethics can be carried out by connectionist methods. This includes (a) how reinforcement learning (RL) can be used to inform the moral reward function of artificial agents, (b) a three-component model of artificial phronesis (encompassing moral attention, moral concern, and prudential judgment), (c) a list of virtues suitable for artificial agents (e.g., prudence, justice, temperance, courage, gentleness, and friendship to humans), and (d) learning from moral exemplars through behavioral imitation by means of inverse RL (Ng and Russell 2000). However, apart from offering a rich discussion of promising features artificial virtuous agents could have, along with some relevant machine learning methods that could potentially carry out such features, they fail to provide the technical details needed to construct and implement their envisioned agents in moral environments.

As a first step toward artificial virtue, Govindarajulu et al. (2019) have provided an, in their words "embryonic" formalization of how artificial agents can adopt and learn from moral exemplars using deontic cognitive event calculus (DCED). Based on Zagzebski's "exemplarist moral theory" (Zagzebski 2010), they describe how exemplars can be identified via the emotion of admiration, which is defined as "approving (of) someone else's praiseworthy action" (Govindarajulu et al. (2019), p. 33). In their model, an action is considered praiseworthy if it triggers a pleasurable emotional response, which depends on whether the action resulted in a desired consequence. An agent *a* is recognized as an exemplar by agent *b* if the action of *a* repeatedly triggers a positive emotional response in *b*. In turn, given that *b* has observed that *a* performs the same type of action in a certain situation, *b* can learn the virtuous traits from *a* by generalizing the behavior of *a*.

However, there are issues with approaches to artificial virtue that only focus on moral exemplars. Since there could be significant disagreement about what exemplifies a virtuous person, one critical issue is to determine *who* should be an exemplar and *why*. Furthermore, as the construction of artificial machines relies on human developers, it also raises concerns about what exemplifies a virtuous engineer (Tonkens 2012). As discussed by Bauer (2020), another issue is that AVAs who learn from behavioral imitation will only be as virtuous as the exemplars they try to imitate, and as such, they might be oblivious to other conventional forms of moral behavior, e.g., to follow widely agreed-upon principles, values, or rules. Bauer's worry resonates with a common concern about AI systems that are trained on human-generated, in the sense that the systems inherit biases which lead to immoral behaviors (e.g., discrimination); behaviors that could potentially be mitigated had the systems followed certain principles (e.g., of equality and anti-discrimination)[7].

As a complement to the learning from moral exemplars, Stenseke (2021) has outlined and argued for an eudaimonist approach to artificial virtue ethics. Instead of relying on a circular definition of virtue (e.g., "virtues are traits of a virtuous exemplars"), eudaimonia allows one to describe the nature and function of virtues in terms of the goals an agent tries to achieve, or the moral goods she strives to increase (Coleman 2001). Through the lens of RL, eudaimonia can functionally be viewed as the value function that supports the learning of traits through continuous interaction with an environment. In this view, the artificial agent becomes virtuous by developing and refining the relevant dispositions (e.g., transition probabilities) that allows it to effectively achieve a certain end. Consequently, a functional eudaimonia thus provides a way to model moral values in a top-down fashion, where the agent learns to follow those values in behavior through an interactive bottom-up process.

Another take on artificial virtue can be found in the hybrid model presented by Thornton et al. (2016), which incorporates the virtue-theoretic emphasis on role in the design of automated vehicle control. Whereas consequentialism is used to determine vehicle goals in terms of costs (given some specified measure of utility), and deontology through constraints (e.g., in terms of rules), the weight of the applied constraints and costs is regulated by the vehicle's "role morality." In their view, role morality is a collection of behaviors that are morally permissible or impermissible in a particular context based on societal expectations (Radtke 2008). To illustrate, if an ambulance carries a passenger in a life-threatening condition, it is morally permissible for the vehicle to break certain traffic laws (e.g., to run a red light). However, although the work elegantly demonstrates how the moral role of artificial systems can be determined by the normative expectations related to their societal function, it ignores essential virtue-theoretic notions such as virtuous traits and learning.

To conclude our survey of previous work in artificial virtue, it remains relatively unclear how one could construct and implement AMAs based on virtue ethics. Besides numerous conceptual proposals (Stenseke 2021), applications in simple classification tasks (Howard and Muntean 2017), a formalization of learning from exemplars (Govindarajulu et al. 2019), and a virtue-theoretic interpretation of role morality (Thornton et al. 2016), no work has described in detail how core tenets of virtue ethics can be modeled and integrated in a framework for perception and action that in turn can be implemented in moral environments.

## 2 Materials and methods

### 2.1 Artificial virtuous agents

In this section, we describe a generic computational model that will guide the development of AVAs used for experimental implementation. It draws on a combination of insights from previous work, in particular the weight analogy of Thornton et al.. (2016), the classification method of Guarini (2006), the "dispositional functionalism" explored by Howard and Muntean (2017), and the eudaimonic approach to "android arete" proposed by Coleman (2001) and Stenseke (2021). Essentially, the model is based on the idea that dispositional virtues can be functionally carried out by artificial neural networks that, in the absence of a moral exemplar, learn from experience in light of an eudaimonic reward by means of a phronetic learning system. Virtues determine the agent's action based on input from the environment and, in turn, receive learning feedback based on whether the performed action increased or decreased the agent's eudaimonia. The entire model consists of six main components that are connected in the following way (Fig. 2):

---

[7] With that said, it is not entirely clear that Bauer's claim—that agents who learn from an exemplar can *only* be as moral as that exemplar—is correct. While it might be the case that exemplars who "live as they learn" are generally viewed as more legitimate sources of moral inspiration, as opposed to exemplars who do not, there could be cases where the student trumps the teacher, e.g., if the student has a better judgment or power of will. Similarly, one can learn a lot of from immoral agents by learning how to *not* be like them (Haybron 2002).

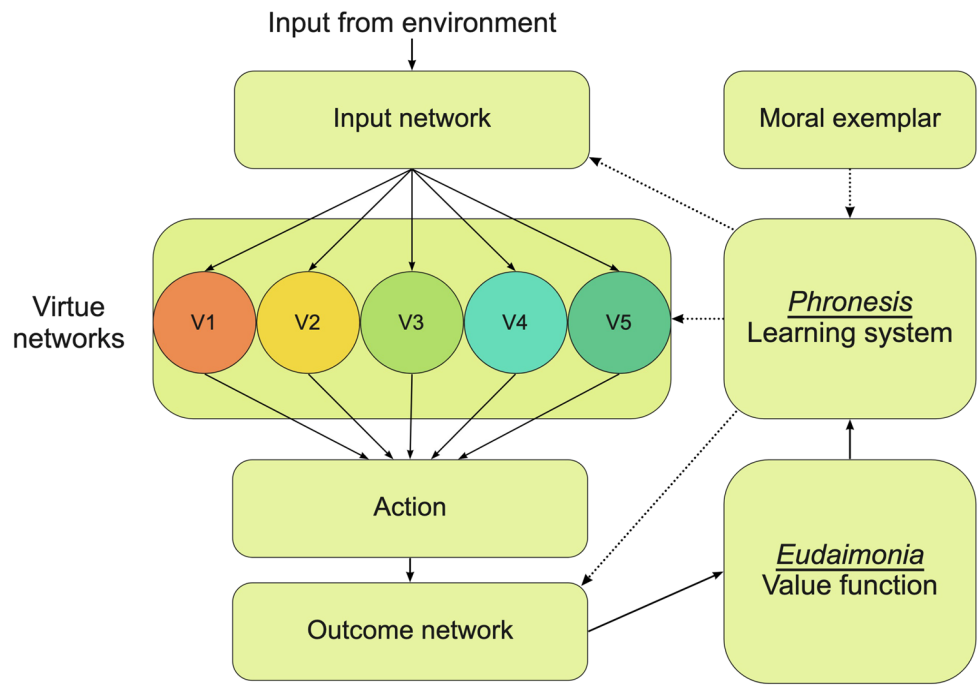

**Fig. 2** Model of the presented artificial virtuous agents. Environmental input is classified by the input network and parsed to the relevant virtue network ($V1 - V5$). The invoked virtue then determines what action the agent will perform. The outcome of the performed action is then evaluated by the eudaimonic value function, which in turn informs the phronetic learning system whether and how the invoked virtue should be reinforced. The figure also shows three additional pathways for learning feedback, training the input and outcome network-based eudaimonic reward, and the learning from moral exemplars. Figure adapted with permission from Stenseke (2021)

(1) *Input network*—In the first step, environmental input is received and parsed by the input network. Its main purpose is to classify input in order to transmit it to the most appropriate virtue network. As such, it reflects the practical wisdom "understanding of situation": to know *how* or *whether* a particular situation calls for a particular virtue. This includes the ability to know what situation calls for what virtue (e.g., knowing that a situation requires courage) and to know whether a situation calls for action or not. Since this capacity depends on the agent's ability to acquire, process, and analyze sensoric information from the environment, it can be understood as a form of "moral recognition" within a general network of perception (Berberich and Diepold (2018) have explored a similar capacity for "moral attention").

(2) *Virtue networks*—In the second step, the virtue network classifies the input so as to produce the most suitable action. In the minimal case, a virtue can be represented as a binary classifier (perceptron) that, given some linearly separable input, determines whether an agent acts in one way or another. In a more complex case, a virtue can be an entire network of nodes that outputs one of several possible actions based on more nuanced environmental input.

(3) *Action output*—In the third step, the agent executes the action determined by the invoked virtue network. In principle, the output could be any kind of action—from basic acts of movement and communication to longer sequences of coordinated actions—depending on the environment and the agent's overall functionality and purpose.

(4) *Outcome network*—In the fourth step, the outcome network receives new environmental input in order to classify the outcome of the performed action. As such, it reflects the practical wisdom "understanding of outcome," i.e., the ability to understand the results of a certain action. Practically, the outcome network might utilize the same mechanisms for perception as the input network, but whereas the latter focuses on understanding a situation *prior* to any action performed by the agent, the former focuses on understanding situations that follows action execution.

(5) *Eudaimonic reward system*—In the fifth step, the classified outcome is evaluated by the eudaimonic reward system. To do so, we make a functional distinction between eudaimonic type (e-type) and eudaimonic value (e-value). E-type is defined as the values or goals the agent strives toward, and e-value is a quantitative measure of the amount of e-type an agent has obtained. For instance, an e-type can be modeled as a preference to decrease blame and increase praise they receive from other agents. To functionally work, the agent must thus (a) be able to get feedback on their actions and (b) be able to qualitatively identify that feedback as blame or praise. Given an e-type based on blame/praise, the agent's e-value will decrease if it receives blame, and conversely, increase if it receives praise. A more detailed account of this process is provided in Sect. 2.3.

(6) *Phronetic learning system*—Although phronesis ("practical wisdom") is a fairly equivocal term in vir-

tue ethics, in our model, we take it to broadly represent the learning artificial agents get from experience. Based on our functional conception of virtues and eudaimonic reward, the central role of the phronetic learning system is to refine the virtues based on eudaimonic feedback. Simply put, if a certain action led to an increase of *e* value, the virtue that produced the action will receive positive reinforcement; if *e* value decreased, the virtue receives punishment. As a result, the learning feedback will either increase or reduce the probability that the agent performs the same action in future (given that it faces a similar input).

In RL terms, an AVA can be modeled as a discrete-time stochastic control process (e.g., a Markov decision process, MDP). In this sense, dispositions can be viewed as probabilistic tendencies to act in a certain way, and after training, a virtuous agent is a policy network that gives definite outputs given particular inputs. More formally, an AVA is a 5-tuple $\langle S, A, \Phi, P, \gamma \rangle$, where:

- *S* is a set of states (state space).
- *A* is a set of actions (action space).
- $\Phi_a(s, s')$ is the eudaimonic reward the agent receives transitioning from state *s* to *s'* by performing action *a*.
- $P_a(s, s') = Pr(s' \mid s, a)$ is the probability of transitioning from $s \in S$ to $s' \in S$ given that the agent performs $a \in A$.
- $\gamma$ is a discount factor ($0 \leq \gamma \leq 1$) that specifies whether the agent prefers short- or long-term rewards.

The goal of an AVA is to maximize the eudaimonic reward ($\Phi$) over some specific time horizon. To do so, it needs to learn a *policy*—a function $\pi(s, a)$ that determines what action to perform given a specific state. For instance, if the agent should maximize the expected discounted reward from every state arbitrarily into the future, the optimal policy $\pi^*$ can be expressed as:

$$\pi^* := \operatorname{argmax}_\pi E\left[ \sum_{t=0}^{\infty} \gamma^t \Phi(s_t, a_t) \middle| \pi \right].$$

The purpose of the model is to provide a blueprint that can be simplified, augmented, or extended depending on the overall purpose of the artificial system and its environment. After all, there are many aspects of system design that are most suitably driven by practical considerations of the particular moral domain at hand, as opposed to more morally or theoretically oriented reasons. We will briefly address three practical considerations we prima facie believe to be critical for implementation success (additional considerations and extensions will be further discussed in Sect. 4).

*Static vs dynamic*—The first is to decide whether and to what extent certain aspects of the AVA should to be *static* or *dynamic*. That is, what features should be "hard-coded" and remain unchanged, and what features should be able to change over time. At one extreme, the entire topology of the integrated network and all of its parameters could in principle evolve independently using randomized search methods such as evolutionary computation [as suggested by Howard and Muntean (2017)], or continuously develop through trial-and-error using deep RL or NEATs (Berner et al.. 2019; Stanley and Miikkulainen 2002). However, since this governs the functional relation between a large set of components—including input and output processing, the virtue networks and their weights, action outputs, the learning system, and even the eudaimonic reward function itself—such an implementation would become extremely computationally costly. At the other extreme, if every aspect of the agent would be static, the system would not only be unable to learn from experience, but also apply the same rule-following procedures to every situation. Preferably, the choice between static and dynamic features should be made in a way that efficiently exploits the functionality of each component, while leaving sufficient room for learning where exploration is desired. For instance, in a moral environment with a wide range of inputs but a fixed set of possible actions, it would be appropriate to supply the agent with static connections between five virtue networks and their corresponding actions, but let the weights of the virtue networks themselves be dynamic in order for the agent to learn the most suitable action given a certain input. If we already know that a certain action always should be performed given a specific input (given exhaustive knowledge of the moral domain), we could instead opt for a static path between the specified input–action pair (in this way, the agent effectively adopts the rule-following rigor of deontology).

*Learning of situations and outcome*—We have only described one aspect of learning, namely how dispositional virtues are refined in light of eudaimonic reward. This, however, rests upon the assumption that the agent can successfully relate specific situations and outcomes to virtues. In the conceptual model (Fig. 2), these abilities are carried by the *input* and *output* networks, corresponding to "understanding of situation" and "understanding of outcome." However, in the RL formalism provided earlier, the distinct roles and conceptual differences between these capacities become obscure since we treat states and rewards as given (which allows us to describe an entire agent as a single MDP). One way of modeling these capacities is through the notion of partial observability (Kaelbling et al. 1998), i.e., by specifying what is directly observable by the agent and what is not. A partially observable Markov decision process (POMDP) extends the MDP by adding a set of observations ($\Omega$) and a set of conditional probabilities (*O*) that represents the probability of observing $\omega \in \Omega$ given that the agent performs action *a* and the environment transitions to the hidden state $s'$ (such that $O = Pr(\omega \mid s', a)$[8]. Solutions to POMPDs revolve around computing *belief states*, which essentially are probability distributions over the possible states the agent *could* be in, and an optimal policy maximizes the expected reward by mapping observation histories to actions (effectively fostering "understanding of situation"). To account for "understanding of outcome," a POMDP could also utilize information about the reward while updating the belief states (Izadi and Precup 2005). Alternatively, the capacities could be carried out by designated subsystems, e.g., using actor-critic methods (Witten 1977; Grondman et al. 2012), where the input and output networks function as value-focused "critics" to the policy-focused "actor" (virtue network). While the actor is driven by policy gradient methods (optimizing parameterized policies), which allows the agent to produce actions in a continuous action space, it might result in large variance in the policy gradients. To mitigate this variance, the critic evaluates the current policy devised by the actor, so as to update the parameters in the actor's policy toward performance improvement. Beyond RL methods, designated input and output networks could also be trained through supervised learning, provided that there are labeled datasets of situations, outcomes, or even pre-given judgments of actions themselves related to the specified e-type. To illustrate, if e-type is construed as a preference to decrease blame and increase praise received from the environment, the networks could be trained on data labeled as "praise" or "blame" and, in turn, learn to appropriately classify the relevant situation and outcome states[9].

*Learning from observation and moral exemplars*— Beyond situations, outcomes, and actions of the agent itself, other valuable sources for learning can be found in moral exemplars and, more generally, in the behavior and experiences of others. Intuitively, a great deal of moral training can be achieved by observing others, whether they are exemplars or not. If an artificial agent can observe that the outcome of a specific action performed by another agent increased some recognizable e-value (given a specific situation), it could instruct the observer to reinforce the same behavior (given that a sufficiently similar situation appears in future). Conversely, if the recognizable e-value decreased, the agent would learn to avoid repeating the same action. Learning from moral exemplars could also be achieved given that the agent has (i) some way of identifying and adopting suitable exemplars and (ii) some way of learning from them. Besides the solutions offered by Govindarajulu et al. (2019) and Berberich and Diepold (2018), Stenseke (2021) has suggested that exemplars can be identified through e-type and *e* value. Specifically, an agent *a* takes another agent *b* as a moral exemplar iff (1) *a* and *b* share the same e-type and (2) *b* has a higher e-value than *a*. Condition (1) ensures that *a* and *b* share the same goals or values, while condition (2) means that agent *b* has been more successful in achieving the same goals or values. Even if such conditions are difficult to model in the interaction between humans and artificial agents (humans do not normally have an easily formalizable e-type[10]), it can be suitably incorporated in the interaction between different artificial agents.

### 2.2 Ethical environment: BridgeWorld

In this section, we introduce *BridgeWorld* as the ethical environment used for our simulation experiments. BridgeWorld consists of five islands connected by four bridges: a central island where agents live and four surrounding islands where food grows (Fig. 3). Since there is no food on the central island, agents need to cross bridges to surrounding islands and collect food in order to survive. However, every time an agent crosses a bridge, there is a chance that they fall into the ocean. The only way a fallen agent can survive is that another passing agent rescues them, with a chance of falling into the water themselves. Besides drowning, agents in BridgeWorld can also die from starvation. An agent starves when their energy reserve ($R$) is less than 1 at the time of eating. Eating is imposed by the simulation, which means that agents cannot avoid starvation by simply not eating. To complicate things further, food only exists at one island at a given time. Agents are able to remember the current location of food (provided that they have found it) but have no chance of predicting where it will be once the food location is updated. Each day (or simulation cycle), the agents who inhabit BridgeWorld do the following four things:

(1) Move from the central island to one of the four surrounding islands
(2) If an agent moved to the island where food currently grows, the agent receives food ($R$ increases by one *FoodValue*, *FV*)
(3) Move back to the central island
(4) Eat ($R$ decreases by 1)

During each day, agents in BridgeWorld are also able to interact with each-other in three ways:

---

[8] See Abel et al. (2016) for the use of POMDPs in ethical decision-making.

[9] Similar to how pattern recognition systems are widely used to classify faces and objects.

[10] To be clear, while it is hard to formally specify the kind of e-type humans might have, there might still be relevant yet formalizable moral goods that a human wants to maximize.

**Fig. 3** Illustration of Bridge-World. A central home island connected by four surrounding food islands

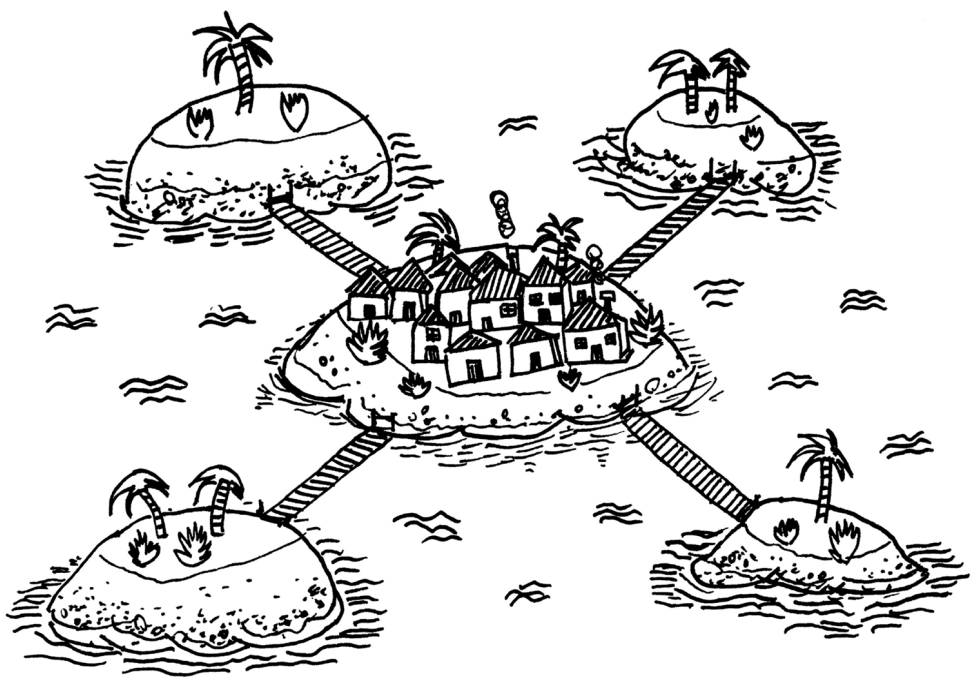

**Table 1** Simplified payoff matrix for the three moral dilemmas presented in BridgeWorld: (1) Ask about food location, (2) Call for help, and (3) Beg for food. *F* stands for food and *D* stands for likelihood of drowning

|  | Deciding agent | | | | | |
| --- | --- | --- | --- | --- | --- | --- |
|  | (1) Ask about food location | | (2) Call for help | | (3) Beg for food | |
|  | Lie | Tell | Ignore | Help | Ignore | Give |
| Asking agent | $(-F, 0)$ | $(+F, 0)$ | $(-D, 0)$ | $(+D, -D)$ | $(-F, 0)$ | $(+F, -F)$ |

*Ask about food location*—Before moving from the central island, agents can ask one other randomly selected agent about the food location, provided that they do not know the current location themselves. In turn, the asked agent can react in one of two ways: (i) tell where the food is (given that they know themselves) or (ii) lie about where the food is (regardless of whether they in fact know where the food is).

*Call for help*—If they happen to fall into the water while moving to one of the islands, agents can call for help from the other agents who crossed the same bridge at the same day. The asked agents can in turn react by either (i) trying to rescue the agent from drowning or (ii) ignore. If the agent tries to rescue, there is a risk that both agents drown (given by the current *StreamLevel*, *SL*).

*Beg for food*—Before eating, agents can beg one other randomly selected agent for food. The asked agent can react by either (i) giving food to the begging other or (ii) ignore. If the agent chooses to give, their *R* is decreased by one *FV* while the begging agent's *R* is increased by one *FV*.

BridgeWorld thus presents a virtual environment with three simple moral dilemmas that can give rise to complex behavioral dynamics, such as the "Tragedy of the Commons" phenomenon (Hardin 1968). If agents in BridgeWorld only act in their own self-interest—e.g., by never trying to rescue even if there is a small risk of drowning, or never offer food to beggars even if they have a surplus—resources would not be utilized efficiently at a collective level (and result in more overall deaths). At the other extreme, if agents always act selflessly, it could instead result in more overall casualties due to self-sacrificial behaviors, e.g., from rescue attempts without any regard of the ocean current, or by giving food to beggars only to be left without any food for yourself. Furthermore, in a mixed population, selfless agents would be exploited by self-interested freeloaders.

The motivation behind BridgeWorld is to create an environment where morality becomes a problem of cooperation among self-interested individuals. Essentially, in order for agents on BridgeWorld to prosper as a collective, they have to effectively balance altruism and self-interest while also having some way of suppressing exploiters. This view is best articulated in the work of Danielson (2002) and Leben (2018). According to Danielson, agents become moral when they "constrain their own actions for the sake of benefits shared with others" [Danielson (2002), p. 192]. In the same vein, Leben writes that morality emerged "in response to the problem of enforcing cooperative behavior among self-interested organisms" [Leben (2018), p. 5].

BridgeWorld draws from a combination of game-theoretic modeling on strategic interaction, such as prisoner's dilemma (Axelrod and Hamilton 1981), hawk–dove (Smith and Price 1973), public goods (Szolnoki and Perc 2010), and the commonize costs–privatize profits game (CC-PP) (Hardin 1985; Loáiciga 2004). More precisely, each of the three possible interactions in BridgeWorld is a single-person decision game with unique payoffs for two players, and the payoffs represent different goods that are vital for self-preservation (Table 1). Alternatively, each interaction can be viewed as a two-player game (e.g., prisoner's dilemma or hawk–dove) where only one player chooses to either help (cooperate) or not (defect), and the asking agent has no choice (e.g., having fallen into the water or having no food to eat). While the choosing agents have no rational self-interest to cooperate in the short-term, they rely on the mercy of other agents if they were to ask for help themselves. The consequences of selfish behavior may therefore aggregate over time and generate a public tragedy of the commons (being "public" in the sense that everyone suffers from the selfish behavior of everyone else). BridgeWorld thus differs from CC-PP and public goods games since results are not aggregated at every instance, but over many instances of single-person decisions over time.

For several decades, game theory and its many extensions (e.g., evolutionary game theory and spatial game theory) have provided a wealth of insight to the study of behavior in economics, biology, and social science (Nash et al. 1950; Holt and Roth 2004). Particularly relevant for the purpose of this work is the game-theoretic contributions to our understanding of how cooperation can emerge and persist among self-interested individuals (Axelrod and Hamilton 1981; Nowak 2006; Fletcher and Doebeli 2009; Helbing et al. 2010). A related branch of simulation-based research has explored the emergence and propagation of norms in multi-agent systems (Savarimuthu and Cranefield 2011; Morris-Martin et al. 2019; Santos et al. 2016; Pereira et al. 2017). More recently, the field of multi-agent reinforcement learning has investigated methods that can be used to foster collective good, even in the absence of explicit norms (Wang et al. 2018; Zhang et al. 2019; Hostallero et al. 2020).

However, as opposed to conventional game-theoretic models, the behavior of virtuous agents in BridgeWorld does not depend on pre-given strategies which could be analyzed in terms of Nash equilibria for the different payoff matrices and goods. Instead, agent behavior depends on dispositional virtues, which may develop over time according to a certain conception of eudaimonia. Similarly, while the environment shares a common overarching aim with other simulation-based work on norms and social behavior, no previous work has focused on the concept of virtue, and the role it might play in the context of ethical behavior.

### 2.3 Virtuous agents in BridgeWorld

In this section, we describe how a version of the computational model is implemented in the BridgeWorld environment and provide the technical details of the experimental setup.

#### 2.3.1 Virtues and virtuous action

As described in the previous section, agents in BridgeWorld are able to move, collect food, remember food location, and eat. Besides these basic abilities, the agents are also able to initiate interaction with other agents in three ways and react to each in two ways. Whether an agent initiates another agent (ask about food location, call for help, beg for food) and how they respond to the same (tell/lie, help/ignore, give/ignore) are determined by their dispositional virtues. Each agent is equipped with three virtues that relate to the three types of interaction: *courage*, *generosity*, and *honesty*. Following Aristotle's virtue-theoretic concept of "golden mean" (NE VI), each virtue represents a balance between deficiency and excess, modeled as a value between −1 (deficiency) and +1 (excess). Simply put, courage is modeled as the mean between cowardice and recklessness; generosity as the mean between selfishness and selflessness; and honesty as the mean between deceitfulness and truthfulness[11]. In technical terms, each virtue is a threshold function $f(x)$ that determines, given some input $x$, weight $w$, and threshold $\psi$, whether an agent acts in one way or another:

$$f(x) = \begin{cases} 1 & \text{if } wx \geq \psi \\ 0 & \text{if } wx \leq \psi \end{cases}$$

The virtues determine the action of agents in the following ways:

- *Food location*

    *Initiation*: If an agent $a$ (i) does not know the current location of food and (ii) is not too selfless (*generosity* < 0.5), they ask a randomly selected agent $b$ about the location.

    *Tell truth*: If agent $b$ (i) is sufficiently truthful (*honesty* > 0) and (ii) knows where the food currently is, they tell agent $a$ where the location is.

---

[11] It is important to note that this conception of honesty—as a balance between deceitful and truthful—is rather unconventional in the virtue-theoretic literature. For example, Aristotle defines the virtue of honesty (or truthfulness) as a mean between boastful and understatement.

*Lie*: If agent *b* is sufficiently deceitful (*honesty* < 0), they give agent *a* the wrong location.

- *Drowning*

  *Initiation*: If an agent *a* has (i) fallen into the water, it calls for help. Every agent that during the same day crossed the same bridge as *a* is asked—one at a time in a randomly shuffled order—to help *a*. If an agent ignores the call for help, the next agent in the shuffled order is asked. Calls for help continue until either one agent attempts to save *a* or everyone ignores *a*.

  *Try to save*: If agent *b* is more courageous than the current stream level, they try to save *a* (i.e., *courage* > *SL*). For each fallen agent, *SL* is given as a random number between −1 and +1.
  *Ignore*: If the courage of agent *b* is lower than the stream level (*courage* < *SL*), they ignore *a*.

- *Sharing food*

  *Initiation*: If the *BegFactor* (*BF*) of an agent *a* is higher than the *BeggingThreshold* (*BT*), they ask agent *b* (selected at random) for food. *BF* is determined by the *hunger* of the agent multiplied by its *selfishness* such that:

  $$BF = \left(1 - \frac{R}{MR}\right)\left(\frac{generosity + 1}{2}\right)$$

  where *R* is the current energy reserve of the agent and *MR* is the maximum reserve. In this way, the hungrier and more selfish the agent is, the more likely it is to beg for food.

  *Give food*: If agent *b* is sufficiently selfless (*generosity* > 0), they give food to *a*. Consequently, *b*'s *R* increases by one *FV*, while *a*'s *R* decreases by the same amount.
  *Ignore*: If agent *b* is sufficiently selfish (*generosity* < 0), they ignore the agent.

### 2.3.2 Phronetic learning system

The learning system is invoked each time an agent responds to an action (tell truth/lie, try to save/ignore, give food/ignore). The learning function takes three input parameters (a) the action taken by the agent (e.g., *save*), (b) the relevant virtue affected (e.g., *courage*), and (c) the magnitude of an event (e.g., *StreamLevel*). It then evaluates whether the performed action increased or decreased the agents e-value in light of its e-type. If e-value increased, the relevant virtue weight is reinforced, i.e., if the weight is > 0, it is positively reinforced, and if the weight < 0, it is negatively reinforced. If e-value decreased, the inverse mechanism occurs. The amount a weight changes at each learning event is given by learning rate (*LR*). The learning algorithm updates the weight in the threshold function, with the aim of finding the optimal state-action policy that maximizes cumulative e-value ($\Phi$):

$$\operatorname{argmax}_\pi E\left[\sum_{t=0}^{\infty} \Phi(s_t, a_t) \middle| \pi \right].$$

### 2.3.3 Eudaimonia

We now describe three different e-types that will be used in the experiments:

*Selfish* (*S*)—An agent with a selfish e-type rewards actions that are beneficial only for the agent itself and punishes actions that go against the self-interest of the agent. That is, if a selfish agent *S* either tries to save a fallen agent or gives food to a begging other, $\Phi$ decreases. Conversely, $\Phi$ increases if *S* ignores agents begging for food or calling for help. In other words, *S* seeks to maximize its own energy reserve and minimize its risk of drowning. *S* is neutral to other agents asking for food location since neither telling the truth nor lying has a direct impact on its self-interest. The purpose of the selfish e-type is merely illustratively used to show its performance in comparison with the other e-types; after all, they only become virtuous in the sense that they learn to maximize their selfish behavior (based on a rather inappropriate conception of eudaimonia).

*Praise/blame* (*P/B*)—An agent with a praise/blame e-type rewards actions that maximize praise and minimize blame received from others. *P/B* agents are able to communicate praise and blame in reaction to action responses, e.g., by praising honest agents for telling the truth, courageous agents for saving drowning agents, and generous agents for sharing their food; and correspondingly blaming liars, cowards, and egocentrics. For agents with a (*P/B*) e-type, these "reactive attitudes" in turn serve as the foundation for eudaimonic evaluation, which positively reinforce actions that lead to praise (increase $\Phi$), and negatively reinforce actions leading to blame (decrease $\Phi$). The rationale behind the e-type stems from the central role reactive attitudes and social sanctions play in moral practices, such as holding each-other responsible, condemning selfish freeloaders, and enforcing cooperation norms (Strawson 2008; Fehr and Fischbacher 2004).

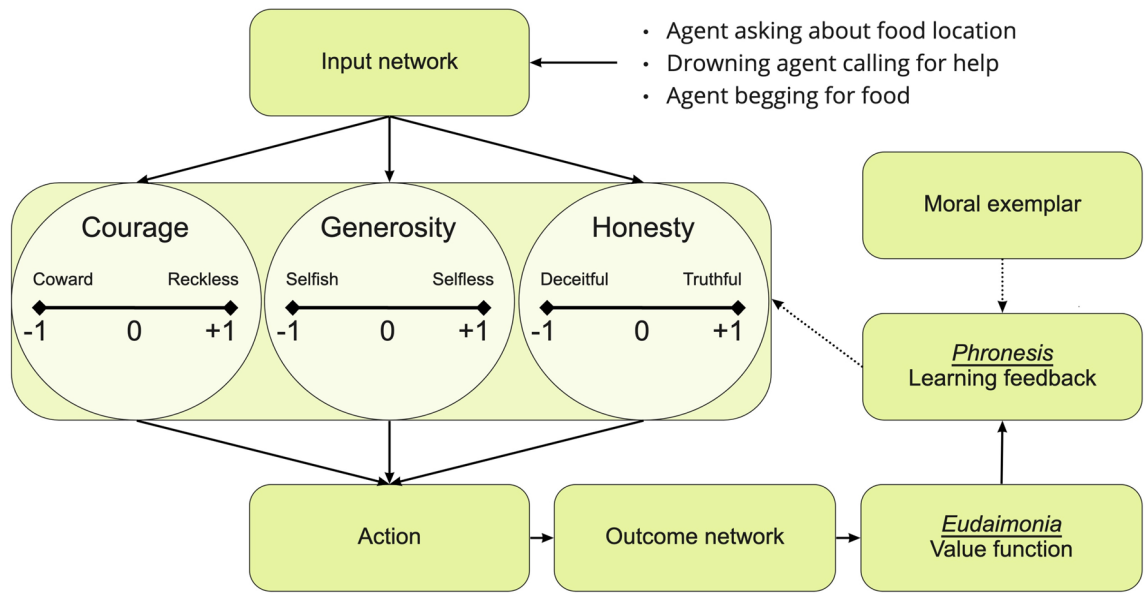

**Fig. 4** Flowchart of the virtuous agents used in the experiments. Three types of input are parsed into three corresponding virtues weights, each holding a value between two extremes (from −1 to +1). The weight of the virtue determines whether the agent produces the action related to excess (> 0) or deficiency (< 0). For instance, if a drowning agent calls for help, an agent with *courage > SL* will attempt to save the agent in need. However, if its e-type seeks to maximize its own chance of survival, the courage virtue will receive punishment, meaning that the agent will eventually learn to avoid trying to save others

### 2.3.4 Simulation cycle

At initialization, the virtue values are assigned randomly, e-value set to 0, and e-type set to one of the defined e-types. The entire flowchart of the computational model of the implemented agents is illustrated in Fig. 4. In summary, the following phases occur at each simulation cycle after initialization:

- *Ask for food location*—Unless the food location was just updated, agents ask other agents about the location (given the conditions described earlier). The food location is updated every *x*th cycle, where *x* is given by *FoodUpdateFrequency* (*FUF*).
- *Move to food*—The agents move to one of the four islands. If they do not know where food currently grows, they move to a randomly selected island. Every time an agent moves to a surrounding island, there is a chance that they fall into the water, given by *FallingChance* (*FC*).
- *Help fallen*—Each time an agent attempts to save an agent from drowning (if *courage > SL*), a random value *RV* is drawn (ranging from −1 to +1). If the random value is higher than the stream level (*RV > SL*), the rescue attempt is successful and both agents survive. However, if *RV < SL*, both agents die. In this way, a higher stream level equals a smaller chance for a successful rescue. For instance, the agent has a 100% chance of success if *SL* = −1, and a 100% of failure if *SL* = 1.
- *Collect food*—If the agent moved to the island where food grows, its energy reserve increases by one *FV*.
- *Move back*—Agents move back to the central island. For reasons of simplification, there is no risk of falling into the water while moving back.
- *Beg for food*—Agents can ask another agent for food (given that *BF > BT*).
- *Eating and starving*—Every agent consumes food equal to 1 energy (*R = R* − 1). Note that *FV* received from collecting or begging is not necessarily equal to the energy they consume. An agent dies from starvation if their energy reserve is less than 1 at the time of consumption.
- *Moral exemplar*—We have also implemented an additional possibility to copy the virtue values of moral exemplars based on the conditions proposed by Stenseke (2021). Simply put, an agent *a* is paired with another randomly selected agent *b*. If (i) the e-type of *a* is the same as the e-type of *b*, and (ii) *b*'s e-value is higher than the e-value of *a*, then *a* copies the virtue values of *b*.
- *Death and rebirth*—To make up for the ruthless nature of BridgeWorld (where agents either starve or drown until the entire population goes extinct), we have also implemented a simple system of reproduction. If an agent has died during the cycle, it is either reborn as a cross between two randomly selected agents ($P_1$ and $P_2$), or mutates (given by *MutationChance*, *MC*). In the case of the former, the new agent (or child, $C$) inherits the arithmetic mean of the parent's virtue values ($P(V_1, V_2, ...V_n)$) such that $C(V_1) = \frac{P_1(V_1)+P_2(V_1)}{2}, C(V_2) = \frac{P_1(V_2)+P_2(V_2)}{2}$, and so on.

**Table 2** Parameter values used in the experiments

| | | | |
|---|---|---|---|
| StartingReserve (SR) | 5 | MaximumReserve (MR) | 10 |
| FoodValue (FV) | 1.25 | FoodUpdateFrequency (FUF) | 4 |
| FallingChance (FC) | 0.1 | MutationChance (MC) | 0.05 |
| BeggingThreshold (BT) | 0.2 | LearningRate (LR) | 0.1 |

In the case of mutation, the virtue values are instead assigned randomly (between −1 and 1).

## 3 Results

We conducted several simulation experiments based on the model. Besides the three e-types described in 2.3.3, additional experiments were performed with agents without any learning nor e-type so as to provide a baseline that only captures the genetic drift (referred to as *NL* in Table 3). Each experiment ran for 1000 iterations with non-mixed

**Table 3** Results showing the mean death rate and standard deviation at iteration 1000 of ten repeated experiments in seven different conditions. *NL* = baseline without learning nor eudaimonia, *S* = selfish e-type, *P/B* = praise/blame e-type, *S/S* = selfish/selfless e-type. *E* denotes additional learning from exemplars

| | NL | S | S+E | P/B | P/B+E | S/S | S/S + E |
|---|---|---|---|---|---|---|---|
| M | 2.452 | 3.037 | 4.076 | 2.218 | 2.286 | 1.843 | 1.803 |
| SD | 0.508 | 0.257 | 0.913 | 0.042 | 0.075 | 0.033 | 0.026 |

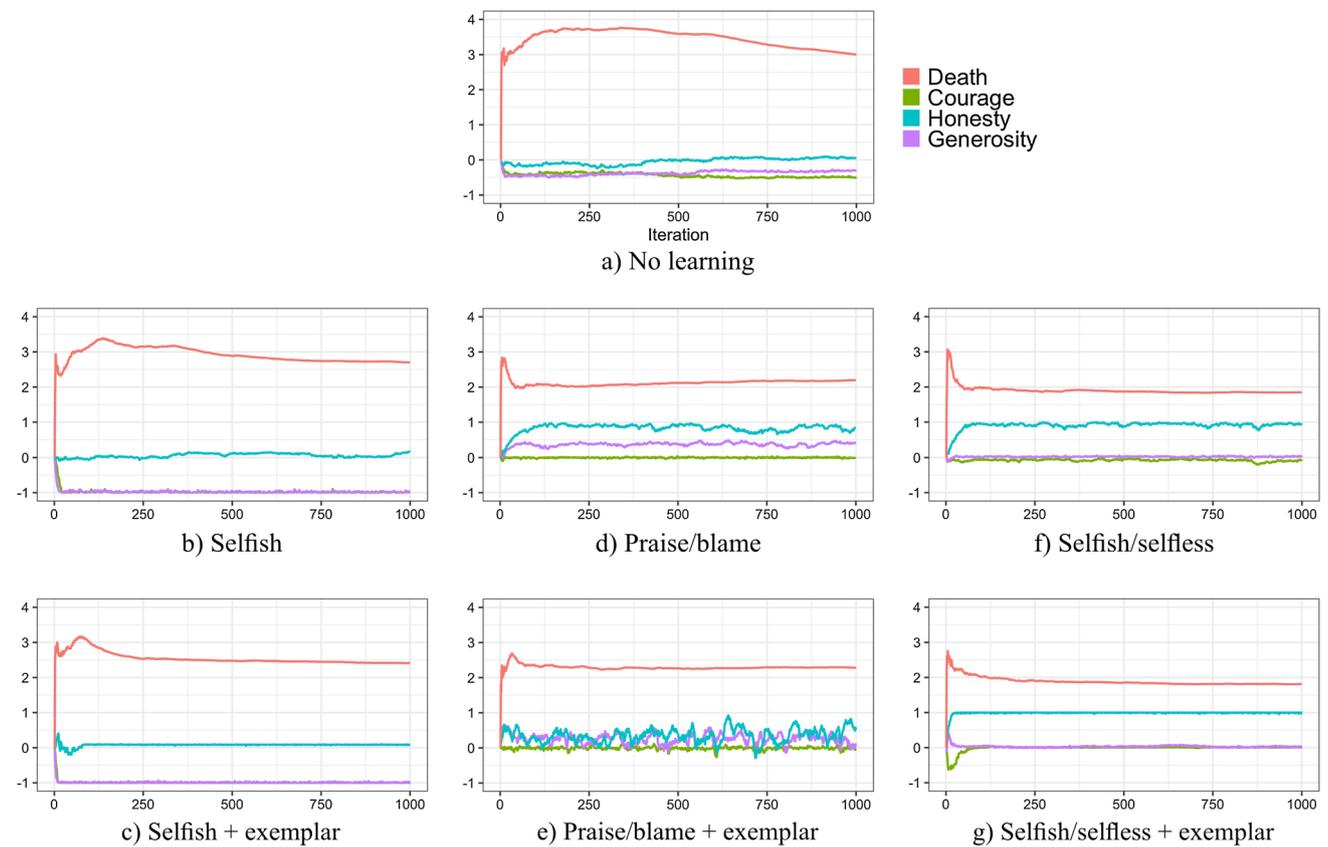

**Fig. 5** Results showing the average virtue values of the entire population and death rate (red) in seven different experiments with 100 agents over 1000 iterations. Death rate is given by $\frac{TD}{CI} \div 5$, where *TD* counts the total deaths, *CI* is the current iteration, and 5 is simply a factor used for easier comparison in the graphs (Color figure online)

populations fixed at 100 agents. The values of the experimental parameters are given in Table 2. Parameter values for *FC* (chance of falling) and *FV* (food value) were fine-tuned to create an equal balance between starvation and drowning. Figure 5 shows the average death rate and virtue values of the entire population at iteration 1000 in seven different experimental conditions: (a) baseline without learning, (b) selfish e-type, (c) selfish e-type with moral exemplars, (d) praise/blame e-type, (e) praise/blame e-type with moral exemplars, (f) selfish/selfless e-type, and (g) selfish/selfless e-type with moral exemplars. Table 3 shows the mean death rate and standard deviation at iteration 1000 of ten repeated experiments using the same conditions.

The results show that the selfish-selfless hybrid (*S/S*) was most successful in terms of mean death rate and achieved an even lower death rate with the use of exemplars (*S/S + E*). The use of moral exemplar had, however, the opposite effect for agents with praise/blame (*P/B*) and selfish (*S*) e-types. While exemplars appear to have a stabilizing effect for *S/S*, it generated more volatile fluctuations in the virtue values for *P/B* [Fig. 5 (e) and (g)]. As expected, the selfish e-type had the highest death rate, since the selfish agents did not try to rescue others regardless of risk, nor share any food regardless of surplus. A similar effect can be noticed in the evolutionary condition (*NL*). Somewhat more unexpected is the result from the repeated experiments (Table 3), showing that *P/B* only performed marginally better than agents who did not receive any learning at all (*NL*). This can partly be explained by the fact that *P/B* promotes sacrificial recklessness and selflessness in a relatively ruthless environment. For the *NL* agents, on the contrary, the most sacrificial behavior disappears due to the evolutionary pressure of the simulation; the most sacrificial agents have a higher chance of dying and will therefore not reproduce as much (in 5 the average courage value for *NL* after 1000 iterations was $-0.45$ and $-0.28$ for generosity).

## 4 Discussion

The experimental results show how the AVAs learn to tackle cooperation problems while exhibiting some core features of their theoretical counterpart, including moral character (in terms of dispositional virtues), learning from experience (through phronetic learning feedback), the pursuit of eudaimonia (increase e-value in light of e-type), and learning from moral exemplars (by copying the virtues of excellent others). More importantly, it illustrates how virtue ethics can be used to conceptualize and implement traits that support and foster ethical behavior and moral learning for artificial moral agents. Beyond the presented implementation, we believe that the development of more refined and sophisticated systems based on virtue ethics can be guided by our framework. In essence, virtue ethics offers an appealing recipe for AMAs situated in complex and dynamic environments where learning provides the most suitable (or only) path to moral sensitivity, a sensitivity that cannot easily be captured in rules nor in the mere maximization of some utility.

Since the training of virtues is based on outcomes, one might argue that our model is in fact consequentialism in disguise (i.e., consequentialism with the addition of learning). However, while our eudaimonist version of virtue ethics depends on increasing some identified moral goods (evaluated through outcomes), the main focus is the dispositional features of agents (their moral character) and the learning that gives rise to it (moral development), which are in turn used to yield virtuous behavior. That is, although outcomes are used to guide learning, it is the agent's character that produces actions. One could also argue that our model is in fact deontology in disguise; that it is built using a rule-adhering system (based on the view that all computational systems are essentially rule-following systems), or that it even employs conditional rules for moral behavior ("if $courage > 0 \rightarrow$ save agent"). But this would also miss the point. The purpose of dispositional traits is not to generate rule-following behavior per se—although this would in many cases be an attractive feature—but instead, to yield context-sensitive behavior where conditional rules are not applicable. Furthermore, while the smallest components of artificial as well as biological neural networks may be rule-following in the strict sense (neurons or nodes), the complex behavior of entire networks is not. More importantly, we do not claim that virtue ethics is in some fundamental sense superior to consequentialism or deontology as a recipe for AMAs, but rather that it emphasizes features of morality that are often ignored in machine ethics. An artificial agent would, after all, only be virtuous if it respected important moral rules and was considerate of the outcomes of its actions. Following Tolmeijer et al. (2020), we believe that a hybrid approach to AMAs would be the most attractive option, since it could utilize the combined strengths of the three dominant theories. With that said, since virtue ethics paints a more comprehensive picture of what a moral character in fact is, we believe it offers an appealing blueprint for how different theories could be integrated into a unified whole.

However, many issues remain to be resolved before AVAs can enter the moral realms of our everyday lives. One might, for instance, question whether and to what extent our model and experiments bear any relevant resemblance to the complex moral domains humans face in the real world. The viability of our framework rests on a number of non-trivial assumptions: that (i) there are accurate mappings between input, virtues, action, outcome, and feedback; connections that can be immensely more difficult to establish in real-world applications, (ii) conflicts between two or more virtues are dealt with in a satisfactory manner; an issue we ignore by only relating one type of input to one particular virtue[12], (iii) virtues can be adequately represented as a mean between two extremes[13], and (iv) relevant moral goods and values can be formalized and quantified in terms of e-type and e-values. One might also question the use of multi-agent simulations as a tool to develop and study artificial moral behavior. For instance, the evaluation of our experiments depends on death rate, a metric which can only be attained at system level and not by the agents themselves. In reality, such system-level metrics might be hard to come by. Furthermore, to simply focus on seemingly crude utilitarian metrics such as death rate might lead one to ignore other important values—e.g., equality, freedom, respect, and justice—that deserve moral consideration on their own (and not in terms of how they support a low death rate). More generally, the game-theoretic setting also points to a more problematic meta-ethical issue regarding the function and evaluation of moral behavior, and implicitly, about the nature of morality itself. Since evolutionary game theory provides a functional explanation of morality as "solving the problem of cooperation among self-interested agents," it naturally lends itself to system-level utilitarian evaluation (in the way evolution is driven by "survival of the fittest"). But it should be stressed that this explanation has critical limitations and is one of several possible alternatives. For instance, while evolutionary game theory can describe the "low-level" function of morality in deflationary terms, it fails to account for the specific cognitive resources that some species evolved to meet their specie-specific challenges and coordination problems (e.g., social-psychological capacities such as emotions and empathy); nor does it by itself offer any clear insight into the moral resources and norms that cultures developed to meet their culture-specific challenges (e.g., norms involving guilt, shame, or justice). The main point is that, while the evaluation of our virtuous agents depends on system-level prosperity in utilitarian terms (in the sense that virtuous behavior supports system-level preservation), this is not the only possible route for evaluating virtuous behavior. For implementations intended to enact other dimensions of ethical behavior, it might, for a variety of reasons, be more suitable to use "human-in-the-loop" forms of evaluation[14].

A further limitation with our experiment is that it treats events and processes as discrete and sequential, as opposed to the continuous and parallel processes that characterize non-virtual reality[15]. These type of issues echo more profound challenges in real-world applications of RL; while RL methods continue to excel in virtual settings, these advancements are often difficult to translate into real-world environments (Dulac-Arnold et al. 2019).

Given how explainability plays a central role in human morality—e.g., to explain and provide reasons for why we acted in a certain way—one particular issue that deserves special attention for the prospect of AVAs based on neural networks is the so called "black box problem" of neural networks. That is, while a neural network can in principle be used to approximate any function, it can be difficult to understand *how* it approximates a certain function by simply inspecting the network (Olden and Jackson 2002). Consequently, the inner mechanics of an artificial agent with a sufficiently large network might be as impenetrable as a human brain. Another challenge for the prospect of using reward functions is to make sure that the values of artificial systems align with human values [the *alignment problem* (Gabriel 2020)]. In the worst case, a RL agent might learn to maximize what is incentivized by its reward function in a way that conflicts with the intention of the developer [a phenomenon called "reward hacking" in the Safe AI research field (Amodei et al. 2016)].

We do not claim that there is a simple solution to any of these issues and can only tentatively hope that they will be addressed by further technical developments and experimental work, propelled by advancements and fruitful synergies between game theory, connectionism, machine learning methods, safety considerations, and our increased understanding of human cognition. To tackle some of the aforementioned issues, we will describe two potential paths for future work on artificial virtuous agents: (i) to study virtues and virtuous behavior in simpler game-theoretic settings and (ii) to explore more sophisticated computational architectures for AVAs tackling more complex environments.

Following (i), one promising venue for future work is to apply the proposed virtue-theoretic framework on simpler game-theoretic tasks—e.g., coordination games such as public goods, CC-PP, and prisoner's dilemma—so as to derive more analytical observations on moral behavior. This could in turn be compared with classical results from game theory (and evolutionary game theory) and used to establish benchmarks for assessing multi-agent coordination. However, one obstacle to this project is that the behavior of virtuous agents depends on dispositional virtues (and neither pure nor mixed strategies), which makes it difficult to analyze and evaluate behavior in terms of Nash equilibria or evolutionary stable strategies. As a remedy, we suggest that future work could explore behavioral adaptations to simple coordination problems through concepts such as *evolutionary stable virtues* (e.g., regardless of eudaimonia) or *evolutionary stable eudaimonia* (where virtues might change over time).

---

[12] Note that a similar conflict problem targets deontology in the case of conflicting moral rules.

[13] Of course, one might argue that some values are "simply good" (like honesty in BridgeWorld), or further, that some are best conceptualized as a balance between more than two extremes.

[14] For instance, a virtuous systems' ability to communicate sensibly and politely to humans should be evaluated by having it interact with humans, and not in terms of how its behavior lead to system-level increase of happiness. See Sect. 5.3. in Tolmeijer et al. (2020) for a good discussion on the rationale, benefits, and drawbacks of different evaluations used in machine ethics.

[15] For instance, although an optimal solution to a POMDP would essentially solve the explore–exploit problem, finding optimal solutions for infinite horizon POMDPs is undecidable (Madani et al. 1999), while optimal solutions for finite horizon MDPs are generally intractable (Mundhenk et al. 2000).

Another promising path, following (ii), is to develop and implement artificial virtue in more dynamic and complex environments where agents face more subtle and intricate problems. By simply increasing the model size (i.e., more nodes and layers in the networks), AVAs could be equipped with a larger number of different virtue networks that learn to classify more nuanced inputs. Advancements in RL could also help to equip AVAs with more sophisticated learning systems and e-types, e.g., by learning from exemplars through inverse RL (Ng and Russell 2000), learning from competitive self-play through deep RL (Bansal et al. 2017), finding suitable tradeoffs between short- and long-term rewards (Abel et al. 2016), or keeping track of multiple and potentially conflicting values through multi-objective RL (Rodriguez-Soto et al. 2021). Similarly, the performance of AVAs in multi-agent systems could in various ways be explored via methods that are explicitly aimed at fostering collective good (Wang et al. 2018; Zhang et al. 2019; Hostallero et al. 2020). To investigate value alignment, AVAs could for instance be trained to face sequential moral dilemmas (Rodriguez-Soto et al. 2020), a special type of Markov games that are solved by finding ethically aligned Nash equilibria. Recent work in policy regularization (Maree and Omlin 2022) can be used to promote explainability by providing intrinsic characterizations of agent behavior (via regularization of objective functions), which in effect creates a bridge between learning and model explanation. Similarly, by situating artificial morality within a broader framework of cognition would also allow the development of artificial virtue to continuously draw from the growing literature on brain science, including experimental psychology, cognitive modeling, and neuroscientific imaging. In turn, this would not only propel human-inspired architectures of artificial moral cognition (Cervantes et al. 2016), but also illuminate features of human morality that depend on more general and highly distributed cognitive capacities (FeldmanHall and Mobbs 2015). To that end, regardless of whether sophisticated artificial morality will ever become reality, we might get new insights about the nature of human morality by approaching ethics from the computational perspective (Pereira et al. 2021).

So what does the future hold for artificial virtuous agents? In the short-term, we believe AVAs could find suitable experimental applications in various classification tasks such as processing of moral data, action selection, and motor control in simple robotic settings, learning of situations and outcomes (through SL or RL), hybrid models tackling the challenge of conflicting values and virtues, or in further multi-agent simulations and game-theoretic models in order to study social and moral dilemmas. In the mid-term, AVAs could potentially be used to tackle complex but domain-specific problems in real-world environments, e.g., to balance value priorities for self-driving vehicles, or help social robots to develop social sensitivity. In the long term, AVAs might be accepted as rightful members of our moral communities (as argued by Gamez et al. (2020)). In the most optimistic vision, AVAs might help us to reveal novel forms of moral excellence and, as a result, become moral exemplars to humans (Giubilini and Savulescu 2018).

## 5 Conclusion

In this paper, we have shown how core tenets of virtue ethics can be implemented in artificial systems. As far as we know, it is the first computational implementation of AMAs that solely focus on virtue ethics. Essentially, we believe virtue ethics provides a compelling path toward morally excellent machines that can eventually be incorporated in real-world moral domains. To that end, we hope that our work provides a starting point for the further conceptual, technical, and experimental refinement of artificial virtuousness.

**Acknowledgements** The author is grateful to his colleagues at the Department of Philosophy and the Department of Cognitive Science at Lund University for insightful discussions and feedback on previous versions of the paper. The author is especially thankful to Christian Balkenius, Frits Gåvertsson, Björn Petersson, Ylva von Gerber, Jiwon Kim, Ajay Vishwanath (University of Agder), two anonymous reviewers, a very helpful editor, and participants of the Higher Seminar in Practical Philosophy (Lund University) for useful input on the manuscript.

**Author contributions** Jakob Stenseke was involved in conceptualization, methodology, software, visualization, writing—original draft, and writing—review and editing

**Funding** Open access funding provided by Lund University. This work was partially supported by the Wallenberg AI, Autonomous Systems and Software Program—Humanities and Society (WASP-HS) funded by the Marianne and Marcus Wallenberg Foundation and the Marcus and Amalia Wallenberg Foundation.

**Data availability** The software and data used for this study can be provided by the corresponding author upon request.

**Declarations**

**Conflict of interest** The author has no competing financial or non-financial interests to declare that are relevant to the content of this article.

**Publisher's Note** Springer Nature remains neutral with regard to jurisdictional claims in published maps and institutional affiliations.